\documentclass[5p,times,procedia]{elsarticle}
\usepackage{ecrc}

\firstpage{1}

\journalname{Procedia CIRP}

\runauth{Lettner et al.}

\makeatletter
\def\ps@pprintTitle{%
  \let\@evenhead\@empty
  \let\@oddhead\@empty
  \let\@evenfoot\@empty
  \let\@oddfoot\@empty}
\def\ps@headings{%
  \def\@oddhead{{\itshape\fontsize{8}{8}\selectfont\hfil\@runauth~/~Accepted manuscript, \@journalname~146 (2026) 421--426\hfil}{\fontsize{8}{8}\selectfont\thepage}}%
  \def\@evenhead{{\fontsize{8}{8}\selectfont\thepage}{\hfil{\itshape\fontsize{8}{8}\selectfont\@runauth~/~Accepted manuscript, \@journalname~146 (2026) 421--426}}\hfil}%
  \let\@evenfoot\@empty
  \let\@oddfoot\@empty}
\gdef\@CopyrightLine{\par\vskip1pc\noindent\textcopyright~2026~The Authors. This is the accepted manuscript of an article published in Procedia CIRP 146 (2026) 421--426, \href{https://doi.org/10.1016/j.procir.2026.03.204}{https://doi.org/10.1016/j.procir.2026.03.204}.\newline
This work is licensed under CC BY-NC-ND 4.0 (\href{https://creativecommons.org/licenses/by-nc-nd/4.0/}{https://creativecommons.org/licenses/by-nc-nd/4.0/})\newline
Peer-review under responsibility of the scientific committee of the 59th CIRP CMS 2026~}
\makeatother

\usepackage{amssymb}

\usepackage[figuresright]{rotating}

\usepackage[bookmarks=false]{hyperref}
\hypersetup{colorlinks,
      linkcolor=blue,
      citecolor=blue,
      urlcolor=blue}

\usepackage{tabularx}
\usepackage{booktabs}
\usepackage{multirow}
\usepackage{subcaption}
\usepackage{stfloats}

\begin{document}
\begin{frontmatter}

\dochead{59th CIRP Conference on Manufacturing Systems 2026 (CMS 2026)}
\title{ScrewSense: Mapping the Invisible -- Anomaly Detection and Error Classification in Screwdriving Processes}

\author[a]{Tobias Lettner \corref{cor1}}
\author[a]{Tatyana Ivanovska}
\author[b]{Anwar Al Assadi}
\author[a]{Christian Bergler}

\address[a]{Department of Electrical Engineering, Media and Computer Science,
Ostbayerische Technische Hochschule Amberg-Weiden, Amberg, Germany}
\address[b]{Fraunhofer Institute for Manufacturing Engineering and Automation, Stuttgart, Germany}

\cortext[cor1]{* \textit{Corresponding author: t.lettner@oth-aw.de}}

\begin{abstract}
The growing volume of process data generated by screwdriving systems, combined with advancements in computing power, data storage, and algorithms, is facilitating the application of artificial intelligence in this field. Accurately detecting and classifying screwdriving errors is crucial for ensuring product quality, avoiding costly rework in manufacturing, and supporting sustainability by facilitating the maintenance, reuse, and recycling of screw-connected products. In this context, machine learning approaches offer a promising alternative to traditional monitoring techniques, such as window-based methods or envelope curves. This study examines the use of deep learning techniques for anomaly detection and error classification in screwdriving processes. The data corpus consists of multivariate time series capturing process parameters, such as torque and rotation angle. For anomaly detection, the embedding space of a vector quantized variational autoencoder was analyzed. In terms of error classification, four different model architectures were evaluated. The results demonstrate the effectiveness of anomaly detection and confirm that deep learning models can distinguish between multiple error categories beyond traditional binary classification. This extends the capabilities of conventional monitoring approaches, achieving accuracies of up to 100\,\%. Additionally, real-world data material was examined in order to further demonstrate model robustness. Overall the results indicate strong generalization capability and highlight the practical potential of the developed models for use in industrial screwdriving processes, eliminating the need for expert knowledge in error identification and significantly reducing manual effort and analysis time.
\end{abstract}

\begin{keyword}
  Anomaly Detection \sep Error Classification \sep Deep Learning \sep Time Series Analysis \sep Industrial Manufacturing \sep Screwdriving Technology
\end{keyword}

\end{frontmatter}

\section{Introduction and Motivation}

Screwdriving systems have evolved into complex technologies that continuously collect process data, such as torque and rotation angle \cite{rabbe2023_schraubtechnik}. This data is primarily used to evaluate the quality of screw joints by determining whether they meet the required specifications (OK versus NOK) \cite{Deprag2020}. Traditionally, this evaluation has relied on monitoring windows that define fixed upper and lower limits for key parameters \cite{Deprag2020, WeberSchraubtechnik2025}.

However, threshold-based methods only assess limited areas of the tightening process. Envelope curves \cite{Schneider1997} can be used to analyze the entire process, but they require extensive parameterization and only indicate whether a process is acceptable or not, without identifying the underlying cause of the deviation.

The trend toward lightweight construction, advanced materials, and higher automation in industrial production is making screwdriving processes increasingly complex. This necessitates adaptive and intelligent monitoring approaches to maintain consistent assembly quality and process reliability \cite{NordLock2021}.

The ongoing rise of artificial intelligence, particularly deep learning, provides innovative solutions to these challenges. They enable flexible and comprehensive analysis of the entire screwdriving process by learning complex patterns and dependencies within the data. Machine learning reduces the need for manual parameter tuning next to expert- and domain-specific rules, supporting higher levels of automation, better quality control, and more reliable processes.

The paper is structured as follows: \nameref{sec:relatedWork} summarizes prior work, \nameref{sec:datasituation} describes the data and preprocessing, \nameref{sec:methodology} details the proposed approach, \nameref{sec:experiments} presents the results, and \nameref{sec:conclusion} summarizes the main findings and outlines future work.

\section{Related Work}
\label{sec:relatedWork}

Recent research has increasingly explored the use of machine learning, particularly modern and state-of-the-art deep learning techniques for monitoring and analyzing the screwdriving process.

Althoefer et al. \cite{althoefer2008automated} used artificial neural networks with radial basis functions to evaluate torque and insertion depth signatures to distinguish between five different error classes. The data originates from computer simulations and experiments involving a rotary torque sensor and an optical shaft encoder with an electric screwdriver. While their model achieved perfect accuracy, it was sensitive to changes in insertion depth.

Meiners et al. \cite{faucris.249327179} evaluated several machine learning models including MLPs, SVMs, and 1D-CNNs, for classifying torque curve anomalies in individual screwdriving operations. The generated dataset contains 228 screwdriving process runs, covering both OK and several distinct failure classes. Using torque data as the only feature, the 1D-CNN achieved the best performance with an accuracy of 86.81\,\%, highlighting the potential of deep learning for process curve-based error classification.

Similarly, Al Assadi et al. \cite{ALASSADI202219} proposed a supervised learning approach to detect stripped and defect-free screw drives using angle-torque profiles from a dismantling process. The custom dataset of over 1,200 samples was used to evaluate a MLP and a 1D-CNN model. The 1D-CNN achieved accuracies between 96\,\% and 98\,\% on unseen test data.

Despite these successes, most approaches focus on predefined scenarios or rely on small, proprietary datasets. This limits their ability to generalize across complex production environments. In contrast, ScrewSense offers a generalizable, deep-learning-based monitoring framework for screwdriving processes, which enables the detection of anomalies and the classification of errors across diverse industrial scenarios, independent of specific system configurations.

\section{Data Situation}
\label{sec:datasituation}

\begin{table*}[!t]
    \renewcommand{\arraystretch}{0.9}
    \footnotesize
    \begin{tabularx}{\textwidth}{Xrrr|Xrr}
        \toprule
        \textbf{Class} & \textbf{assm\_a} & \textbf{dism\_a} & \textbf{reasm\_a}
        & \textbf{Class} & \textbf{dism\_b} & \textbf{dism\_c} \\
        \midrule
        OK & 58 & 59 & 59 & OK & -- & 50 \\
        BURSTING\_TUBE & 48 & 48 & 46 & REFERENCE & 485 & -- \\
        TIGHT\_SCREW & -- & 50 & 50 & REFERENCE\_LARGE & 372 & -- \\
        CROSS\_THREADING & -- & -- & 47 & INCORRECT\_DRIVE & 480 & 101 \\
        DAMAGED\_SCREW\_DRIVE & -- & 51 & -- & TIGHT\_SCREW & 46 & 50 \\
        SCREW\_EXTRACTION & 51 & 53 & -- & DAMAGED\_HEAD\_LARGE & 381 & -- \\
        PLASTIFICATION\_THREAD & -- & 50 & 49 & DAMAGED\_DRIVE & -- & 100 \\
        \bottomrule
    \end{tabularx}
    \caption{Class distribution of all datasets used in this study, showing the available classes and the corresponding number of samples per dataset. The dataset identifiers follow the naming convention \textit{assm} (assembly), \textit{dism} (dismantling), and \textit{reasm} (reassembly), with suffixes \textit{a}, \textit{b}, and \textit{c} denoting different system configurations.}
    \label{tab:dataset_combined}
\end{table*}

\subsection{Dataset Material}
This study utilizes a comprehensive collection of five different datasets on screwdriving processes for assembly, dismantling, and reassembly under various system configurations. The data material provides a diverse foundation for evaluating adaptive and intelligent deep learning driven monitoring approaches. All datasets are publicly available at \url{https://github.com/lettnert/ScrewSense.git}.

The first three datasets, referred to as \textit{assm\_a}, \textit{dism\_a}, and \textit{reasm\_a}, were generated using a DEPRAG 311ER27-0020 screwdriver with a vertical guidance system that was operated manually. Self-tapping Ejot Delta PT screws (3×10 mm) were driven into PA6.6-GF30 plastic specimens. These datasets represent the assembly, dismantling, and reassembly screwdriving operation of identical components. Two additional datasets, \textit{dism\_b} and \textit{dism\_c}, were recorded using different screwdriving systems. The data corpus \textit{dism\_b} was generated with a DEPRAG 326E27-0080-H screwdriver mounted on a UR10e robotic arm, while \textit{dism\_c} was created using a Bosch Rexroth Nexo screwdriving system with a manually operated gantry setup. Both datasets represent dismantling operations. The corresponding class distributions are summarized in Table~\ref{tab:dataset_combined}.

Using the same configuration as for \textit{dism\_b}, an additional set of 36 \textit{real-world data} samples was acquired by loosening screws from lawn mower housings. This data corpus does not contain any class labels.

\subsection{Data Preprocessing}
The first step of the proposed data preprocessing pipeline describes a resampling step, followed by a min-max normalization. Resampling the input signals to a fixed length using linear interpolation~\cite{pytorchInterpolation} ensures consistent input dimensions across all samples. This process can also act as a form of compression or stretching, depending on the original length of the time series. Min-max normalization scales the data to a uniform range~\cite{10.1007/978-3-031-11432-8_8, wu2018group}. This step stabilizes the training process and facilitates faster convergence by mitigating numerical instabilities. Each feature is normalized independently, preserving the relative differences and distributions of the original data, which enhances the robustness and overall performance of the model.

Several data augmentations were applied~\cite{Wen_2021} in order to further improve model generalization, comprising window warping, Gaussian noise injection, spike injection, step-like trend injection, and slope-like trend injection. Each transformation introduces controlled variations with respect to the input signals, enhancing robustness while preserving the underlying temporal characteristics. To avoid distorting the original signal characteristics, only a single augmentation is applied per sample, selected randomly during training.

\begin{figure*}[!t]
    \centering
    \includegraphics[width=0.90\linewidth]{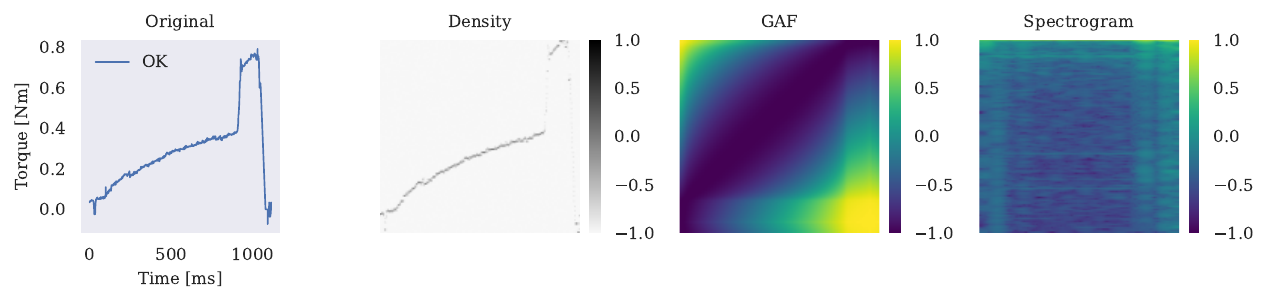}
    \caption{Example of image conversion results. From left to right: original input, density, GAF, and spectrogram representations of the torque signal. Each image has a size of 224×224 pixel. The model uses three-channel inputs, with torque, rotation angle, and time as features. Density images use a 224×224 bin size; GAF is generated with the summation method; spectrogram parameters are nfft = 512, hop\_length = 32, and a Hamming window.}
    \label{fig:conversion_results}
\end{figure*}

In case of image-based models, time series data are converted into image representations using three methods: density transformation via two-dimensional histograms~\cite{numpy2dhistogram}, Gramian Angular Field (GAF) transformation~\cite{inproceedingsEncodingTimeSeries}, and spectrogram transformation~\cite{pytorchSTFT}. Before generating a spectrogram, the mean is subtracted from each sample to remove the DC component and align the signal around zero. The resulting amplitude values are then converted to the decibel scale to better capture signal dynamics~\cite{BerglerDissertation}. Figure~\ref{fig:conversion_results} visualizes the results of all three image conversion techniques.

\section{Methodology}
\label{sec:methodology}

\subsection{Anomaly Detection}
\label{subsec:methodology_anomaly_detection}
In ScrewSense, anomaly detection is performed using a vector quantized variational autoencoder (VQVAE), originally introduced by Oord et al.~\cite{oord2018neuraldiscreterepresentationlearning}. In order to perform a task-specific model adaption, the VQVAE was pre-trained on a subset of the \mbox{ImageNet} dataset~\cite{ILSVRC15}, which enabled it to learn general visual features and robust latent representations. Subsequently, the model was fine-tuned on the target corpus, where the time-series data were converted into two-dimensional images through GAF representations. Fine-tuning was conducted solely on samples labeled as OK or REFERENCE, representing the normal class. After training, each sample was processed through the encoder to obtain its embedding representation. To enable visualization and further analysis, the embeddings were reduced to two dimensions using t-SNE~\cite{van2008visualizing} (e.g. Figure~\ref{fig:experiments_anomalyDetection_results_hidden_states_depragDismantling}). The resulting embeddings were plotted as a two-dimensional  scatter plot, with each sample colored according to its class label. This provides an intuitive view of the clustering structure and potential anomalies.

\subsection{Error Classification}

In the error classification task, four distinct model architectures are utilized, each describing a different deep learning paradigm: MLP-, CNN-, transformer-, and vision-based models. The purpose of this comparison is to assess how effectively each architecture captures temporal dependencies and discriminates among error scenarios in screwdriving processes.

The first model, \textit{PatchTSMixer (PTSM)}~\cite{Ekambaram_2023}, is a lightweight architecture specifically designed for multivariate time series classification. It replaces the self-attention mechanism of transformers with a pure MLP-based design, thereby reducing memory and computational demands while maintaining strong performance on long sequences~\cite{wen2023transformerstimeseriessurvey}. \textit{ConvTimeNet (CTN)}~\cite{cheng2024convtimenetdeephierarchicalfully} represents a CNN-based approach and employs deformable patch embeddings combined with hierarchical convolutional blocks to extract local patterns and capture multi-scale temporal dependencies within the data. \textit{ConvTran (CTR)}~\cite{islam2023comprehensivesurveyapplicationstransformers}, as a transformer-based model, combines convolutional feature extraction with advanced position encoding mechanisms to effectively capture temporal dependencies in time series data. Finally, the \textit{Vision-Transformer (ViT)}~\cite{dosovitskiy2021imageworth16x16words} processes time series data that have been transformed into 2D-image representations. By leveraging the attention mechanism, originally developed for sequence modeling in natural language processing~\cite{NIPS2017_3f5ee243}, the concept was transferred to the image domain by feeding sub-image patches as sequential data to the model, allowing to learn robust and discriminative features for classification.

All models are trained supervised using the provided class labels and the Adam optimizer with exponential learning rate decay. Class imbalance is handled through weighted cross-entropy loss. A 70/15/15 train/validation/test split with early stopping is employed, followed by 5-fold cross-validation, whose scores are reported in the experimental section.

\section{Experiments, Results, and Discussion}
\label{sec:experiments}

\subsection{Experiment-1: Anomaly Detection}
\label{ubsec:experiments_results_exp1}
\begin{figure}[h!]
    \centering
    \includegraphics[width=0.80\linewidth]{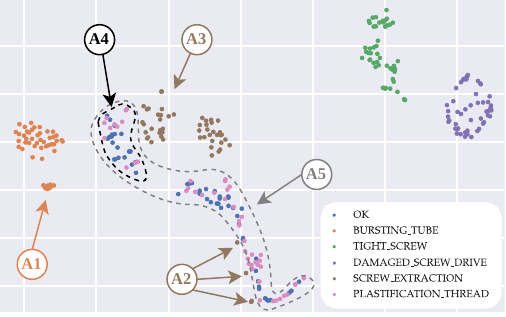}
    \caption{Embedding space of the \textit{dism\_a} dataset. Each point represents a screwdriving sample described by torque, rotation angle, and time, with colors indicating class labels. Multiple anomalies occurred and are highlighted as A1\,-\,A5.}
    \label{fig:experiments_anomalyDetection_results_hidden_states_depragDismantling}
\end{figure}

The first experiment was conducted to analyze the embedding space generated from the datasets. Figure \ref{fig:experiments_anomalyDetection_results_hidden_states_depragDismantling} visualizes the hidden embedding space of the data \textit{dism\_a} corpus.

While some classes are clearly distinct from others, five anomalies were observed and analyzed in more detail. These anomalies are denoted as \textit{A1} through \textit{A5}. Samples belonging to anomaly \textit{A1} exhibit a screwdriving time approximately 140\,ms longer than the other samples. Samples belonging to anomaly \textit{A2} show a breakaway torque that occurs about 50\,° later than in the normal. Anomalies \textit{A3} and \textit{A4} display an earlier breakaway torque of about 25\,°, which also explains their close proximity within the embedding representation. Anomaly \textit{A5} represents an overlap between two classes, where the corresponding samples show a high degree of similarity. Due to the fact that all anomalies can occur in real-world scenarios, no samples were excluded for the adapted error classification experiment.

\begin{figure}[h!]
    \centering
    \includegraphics[width=0.8\linewidth]{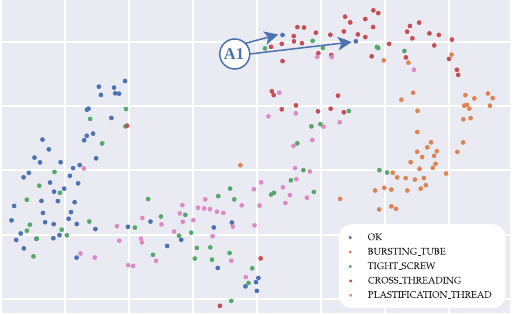}
    \caption{Embedding space of the \textit{reasm\_a} dataset. Each point represents a screwdriving sample described by torque, rotation angle, and time, with colors indicating class labels. Multiple anomalies occurred. As an example, two \textit{OK} samples are highlighted within the cluster dominated by \textit{CROSS\_THREADING} instances.}
    \label{fig:experiments_anomalyDetection_results_hidden_states_depragReassembly}
\end{figure}

Figure~\ref{fig:experiments_anomalyDetection_results_hidden_states_depragReassembly} illustrates the embedding space of the most challenging dataset, \textit{reasm\_a}. As an example, two anomaly samples from the \textit{OK} class were selected, which are located within the \textit{CROSS\_THREADING} cluster.

\begin{figure}[h!]
    \centering
    \includegraphics[width=0.92\linewidth]{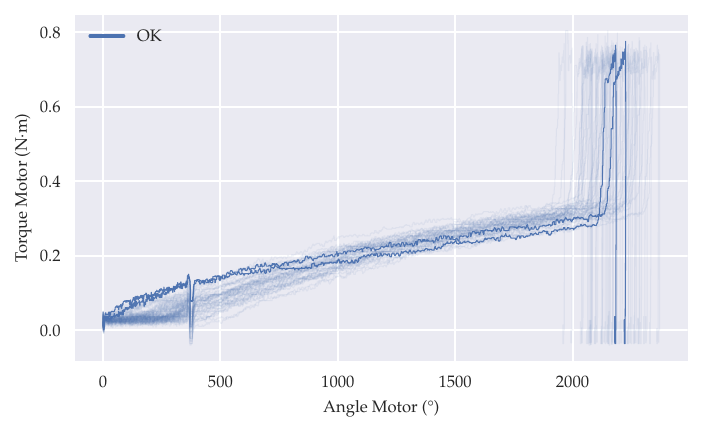}
    \caption{Curve overlay of the highlighted anomaly samples from Figure~\ref{fig:experiments_anomalyDetection_results_hidden_states_depragReassembly}, compared to the normal class samples. The anomaly samples, displayed with higher opacity, exhibit typical cross-threading behavior, characterized by a higher friction torque during the initial rotations.}
    \label{fig:experiments_anomalyDetection_depragReassembly_curveOverlay}
\end{figure}

Figure~\ref{fig:experiments_anomalyDetection_depragReassembly_curveOverlay} visualizes the curve overlay of these samples compared to the normal class samples. These samples exhibit a cross-threading behavior, where the friction torque at the beginning of the screwdriving process is higher than average. This behavior is likely due to mislabeling during data collection. These samples were reassigned to the correct class for the adapted classification experiment.

Similar patterns and anomalies were observed in the datasets \textit{assm\_a}, \textit{dism\_b}, and \textit{dism\_c}. The same analysis procedure and methodology were applied to these three.

\subsection{Experiment-2: Error Classification}
\label{ubsec:experiments_results_exp2}
The second experiment focused on using the proposed models to evaluate their classification performance on each dataset. For evaluation, balanced accuracy (BAC) and F1-score were used. The results are summarized in Table \ref{tab:exp2_results_errorClassification}, with only the BAC reported for simplicity since the F1-score results were similar. All models demonstrate comparable overall performance. PatchTSMixer and ConvTimeNet perform well on simpler datasets. The ViT model with the GAF transformation outperforms the other models on more complex datasets.

\begin{table}[h!]
    \footnotesize
    \resizebox{\linewidth}{!}{%
    \begin{tabular}{lccccc|c}
        \toprule
        \multirow{2}{*}{\textbf{Model}} & \multicolumn{5}{c|}{\textbf{Dataset performance - BAC}} & \multirow{2}{*}{\textbf{Mean}} \\
        \cmidrule{2-6}
        & assm\_a & dism\_a & reasm\_a & dism\_b & dism\_c & \\
        \midrule
        PTSM & \textbf{1.0000} & 0.9157 & 0.7015 & 0.9912 & 0.9602 & 0.9137 \\
        CTN & \textbf{1.0000} & 0.9425 & 0.5362 & \textbf{1.0000} & 0.9701 & 0.8898 \\
        CTR & 0.9889 & 0.9257 & 0.6735 & 0.9867 & 0.9625 & 0.9074 \\
        ViT-Dens. & 0.9944 & 0.9595 & 0.7391 & 0.9944 & 0.9700 & 0.9315 \\
        ViT-Spec. & 0.9944 & 0.9276 & 0.6926 & 0.9532 & \textbf{0.9800} & 0.9096 \\
        ViT-GAF & 0.9944 & \textbf{0.9757} & \textbf{0.7812} & 0.9934 & \textbf{0.9800} & \textbf{0.9450} \\
        \bottomrule
    \end{tabular}%
    }
    \caption{Model classification performance in terms of BAC across datasets, with the right column showing the averaged mean performance across all datasets. The best score for each dataset is highlighted in bold. ViT-GAF achieves the highest mean performance.}

    \label{tab:exp2_results_errorClassification}
\end{table}

For a better comparison, the mean performance of the models over all datasets was calculated using unweighted averaging, treating each dataset as equally important. ViT-GAF achieves the best mean performance with a BAC-score of 0.9450 across all datasets. Therefore, the ViT-GAF will be used for \textit{Experiment-3/4}. ViT-Density shows also strong average performance but does not outperform any other model on a single dataset. The other models show consistently lower average performance.

\subsection{Experiment-3: Adapted Error Classification}
\label{ubsec:experiments_results_exp3}

An additional error classification was conducted using the ViT-GAF model on the adapted datasets obtained from \textit{Experiment-1}. The comparison to the classification results on the original datasets from \textit{Experiment-2} are listed in Table~\ref{tab:error_classification_adapted}.

\begin{table}[h!]
    \footnotesize
    \begin{tabularx}{1\linewidth}{Xcccc}
        \toprule
        \textbf{Model: ViT-GAF} & \multicolumn{2}{c}{\textbf{Original dataset}} & \multicolumn{2}{c}{\textbf{Adapted dataset}}\\
        \cmidrule{2-5}
        \textbf{Dataset} & BAC & F1-Score & BAC & F1-Score \\
        \midrule
        assm\_a & 0.9944 & 0.9937 & \textbf{1.0000} & \textbf{1.0000} \\
        dism\_a & \textbf{0.9757} & \textbf{0.9741} & \textbf{0.9757} & \textbf{0.9741} \\
        reasm\_a & 0.7812 & 0.7661 & \textbf{0.9234} & \textbf{0.9237} \\
        dism\_b & 0.9934 & 0.9982 & \textbf{0.9989} & \textbf{0.9989} \\
        dism\_c & 0.9800 & 0.9799  & \textbf{0.9900} & \textbf{0.9922} \\
        \bottomrule
    \end{tabularx}
    \caption{Error classification results of the ViT-GAF model on adapted compared to original datasets, showing improvements for all datasets except \textit{dism\_a}, which required no adaptation. Better value is highlighted in bold.}
    \label{tab:error_classification_adapted}
\end{table}

\begin{figure}[h!]
    \centering
    \includegraphics[width=\linewidth]{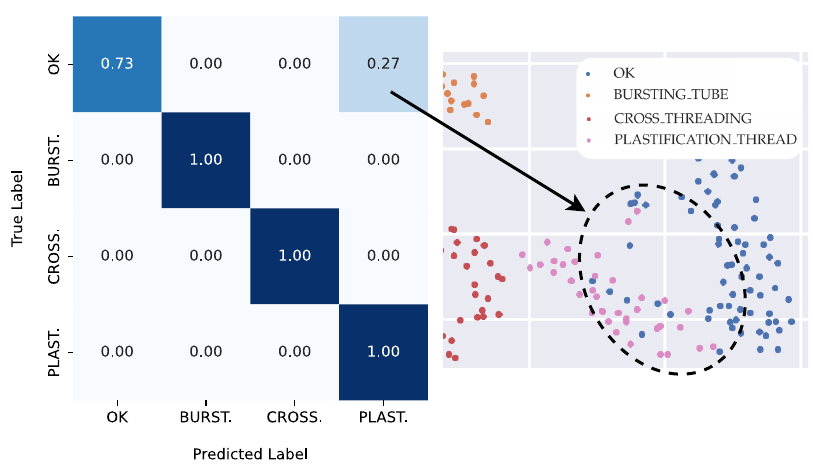}
    \caption{Confusion matrix of the classification result and a subregion of the embedding space of the adapted \textit{reasm\_a} dataset. Misclassifications mainly occur between \textit{OK} and \textit{PLASTIFICATION\_THREAD} class due to their similarity.}
    \label{fig:exp3_confusion_matrix}
\end{figure}

The classification performance improved across all datasets, except of \textit{dism\_a}, because there was no adaption necessary. There was a significant improvement for the most challenging dataset, \textit{reasm\_a}, but some misclassifications remained. The confusion matrix and the embedding space (see Figure~\ref{fig:exp3_confusion_matrix}) reveal the reason for the misclassifications. Samples from the \textit{OK} class were classified as \textit{PLASTIFICATION\_THREAD} due to the similarity between these classes, caused by insufficient manipulation of the test specimens. The embedding space shows an overlap between these two classes.

\subsection{Experiment-4: Generalization to Real-World Data}
\label{ubsec:experiments_results_exp4}

\begin{figure}[h!]
  \centering
  \includegraphics[width=.8\linewidth]{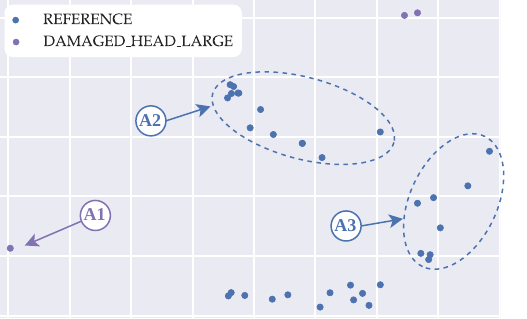}
  \caption{Embedding space of the classified lawn mower data with highlighted anomalies. Sample colors indicate the predicted class labels.}
  \label{fig:embedding_space_lawn_mower}
\end{figure}

The fourth experiment involved a generalization test using real-world data. The ViT-GAF model, trained solely on the \textit{dism\_b} dataset, was applied unchanged to the real-world data. A total of 33 samples were classified as \textit{REFERENCE}, and three as \textit{DAMAGED\_HEAD\_LARGE}. Figure \ref{fig:embedding_space_lawn_mower} illustrates the embedding space, which shows three distinct clusters for the \textit{REFERENCE} samples and one \textit{DAMAGED\_HEAD\_LARGE} sample that appears particularly far away from the other two.

\begin{figure}[h!]
    \centering
    \includegraphics[width=0.75\linewidth]{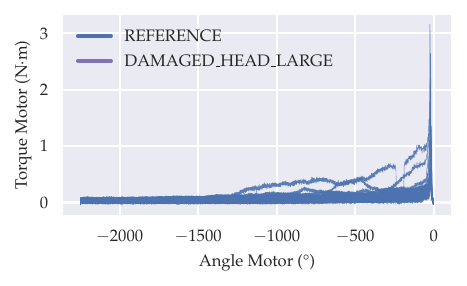}
    \includegraphics[width=0.75\linewidth]{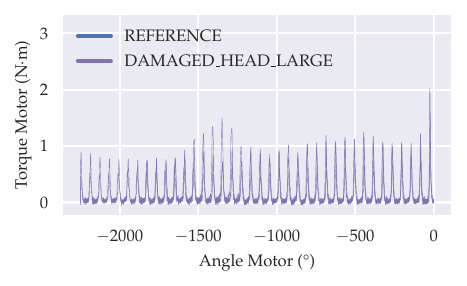}
    \caption{Comparison of classified samples from lawn mower data. The upper subfigure shows the curve overlay of samples classified as \textit{REFERENCE}, and the lower subfigure shows the curve overlay of samples classified as \textit{DAMAGED\_HEAD\_LARGE}.}
    \label{fig:comparison_classified_samples}
\end{figure}

The screwdriving curves, visualized in Figure \ref{fig:comparison_classified_samples}, generally show accurate classification of all samples based on visual inspection. The DAMAGED\_HEAD\_LARGE samples exhibit the typical behavior of a drive slipping over the screw head, resulting in torque peaks during the screwdriving process.

The isolated sample, denoted as anomaly \textit{A1}, exhibits a higher final angle compared to the two class samples, which explains its deviation in the embedding space. This deviation may be caused by an interruption during the screwdriving process. Anomaly \textit{A2} is characterized by a lower breakaway torque relative to the other samples, while anomaly \textit{A3} shows a higher friction torque during the loosening process. Both anomalies can occur in real applications. These anomalies occur within the \textit{REFERENCE} class and can also occur in real applications. A visual inspection of the screwdriving curves shows that the screws are fully loosened and are classified correctly. Each curve follows the characteristic pattern of the loosening process: an initial increase in torque up to the breakaway point, followed by a decrease in torque and continued rotation.

\section{Conclusion and Future Work}
\label{sec:conclusion}

Deep learning techniques can detect anomalies and classify errors in the evaluation of screwdriving processes. The embedding space of the VQVAE model provides a clear visualization of different anomalies, offering a promising approach for analyzing multiple screwdriving curves. The ViT-GAF performed well in error classification, and both models generalized well with respect to real-world production data. One limitation is the reliance on labeled data. Around 50 samples per class were sufficient for good performance. However, differences in system- or recording-specific signals may restrict model transferability across datasets. A direct comparison to existing approaches was not feasible, as no publicly evaluated benchmarks exist for the specific datasets and error types investigated in this work.

Future work includes generating synthetic data to augment training datasets, applying foundation models such as TOTEM~\cite{talukder2024totem}, or leveraging meta-transfer learning techniques \cite{10061349} to improve model generalization. Further advances may include integrating explainable artificial intelligence methods to increase transparency in anomaly detection. For example, anomalies can be visualized directly in time series using models like \mbox{PredTrAD}~\cite{schuster25_interspeech}. Another interesting area of exploration would be detecting errors during the screwdriving process by processing the time series sequentially rather than classifying only the final result. This would enable real-time anomaly detection and early intervention. Additionally, the PyScrew dataset \cite{west2025pyscrewcomprehensivedatasetcollection}, with over 34,000 screwdriving operations and 27 error types, offers a valuable resource for model pre-training and validation. Leveraging this dataset enhances generalization and enable more robust evaluation across diverse screwdriving scenarios, and would also enable future comparisons with other approaches once common benchmarks become available.

\section*{Acknowledgments}
This work was supported by the German Federal Ministry for the Environment, Nature Conservation, Nuclear Safety and Consumer Protection under \emph{Desire4Electronics}.

\bibliography{literature.bib}

@techreport{Deprag2020,
  author       = {{DEPRAG SCHULZ GMBH u. CO. KG}},
  title        = {{Technical Information: Screwdriving Technology and Quality Assurance}},
  institution  = {{DEPRAG SCHULZ GMBH u. CO. KG}},
  year         = {2020},
  url          = {https://www.deprag.com/fileadmin/bilder_content/emedia/broschueren_pics/Allgemein_PDF/D0080/D0080en.pdf},
}

@incollection{Schneider1997,
  author    = {Schneider, W.-D.},
  title     = {{H{\"u}llkurvenverfahren zur Proze{\ss}{\"u}berwachung}},
  booktitle = {Flexibel automatisiertes Taumelnieten},
  publisher = {Springer Berlin Heidelberg},
  address   = {Berlin, Heidelberg},
  year      = {1997},
  pages     = {58--65},
  doi       = {10.1007/978-3-662-07247-9_7}
}

@online{NordLock2021,
  author  = {{Nord-Lock Group}},
  title   = {{Benefits of Using Bolted Joints in the Industrial Sector}},
  year    = {2021},
  url     = {https://www.nord-lock.com/learnings/bolting-tips/2021/industrial-bolted-joints/},
}

@incollection{rabbe2023_schraubtechnik,
  author    = {Niels Rabbe},
  title     = {{Messtechnik in der Schraubtechnik}},
  booktitle = {Handbuch Messtechnik in der industriellen Produktion: Valide Messergebnisse planen, erhalten, auswerten und verteilen},
  editor    = {Robert Schmitt and Edgar Dietrich},
  edition   = {1},
  publisher = {Carl Hanser Verlag GmbH \& Co. KG},
  address   = {München},
  year      = {2023},
  pages     = {359--374},
  isbn      = {978-3-446-46559-6},
  eisbn     = {978-3-446-46626-5},
  language  = {German}
}

@article{althoefer2008automated,
  title     = {{Automated failure classification for assembly with self-tapping threaded fastenings using artificial neural networks}},
  author    = {Althoefer, Klaus and Lara, B and Zweiri, Y and Seneviratne, Lakmal D},
  journal   = {Proceedings of the Institution of Mechanical Engineers, Part C: Journal of Mechanical Engineering Science},
  volume    = {222},
  number    = {6},
  pages     = {1081--1095},
  year      = {2008},
  publisher = {SAGE Publications},
  doi       = {10.1243/09544062JMES546}
}

@article{faucris.249327179,
  author         = {Meiners, Moritz and Mayr, Andreas and Franke, Jörg},
  doi            = {10.1016/j.procir.2020.05.220},
  faupublication = {yes},
  journal        = {Procedia CIRP},
  pages          = {166-171},
  peerreviewed   = {Yes},
  title          = {{Process curve analysis with machine learning on the example of screw fastening and press-in processes}},
  volume         = {97},
  year           = {2021}
}

@article{ALASSADI202219,
  title    = {{Machine learning based screw drive state detection for unfastening screw connections}},
  journal  = {Journal of Manufacturing Systems},
  volume   = {65},
  pages    = {19-32},
  year     = {2022},
  issn     = {0278-6125},
  doi      = {https://doi.org/10.1016/j.jmsy.2022.07.013},
  author   = {{Anwar {Al Assadi} and David Holtz and Frank Nägele and Christof Nitsche and Werner Kraus and Marco F. Huber}}
}

@misc{pytorchInterpolation,
  author = {{pytorch.org}},
  title = {{torch.nn.functional.interpolate}},
  url = {https://pytorch.org/docs/stable/generated/torch.nn.functional.interpolate.html},
  year = {},
}

@InProceedings{10.1007/978-3-031-11432-8_8,
  author="Asesh, Aishwarya",
  editor="Biele, Cezary
  and Kacprzyk, Janusz
  and Kope{\'{c}}, Wies{\l}aw
  and Owsi{\'{n}}ski, Jan W.
  and Romanowski, Andrzej
  and Sikorski, Marcin",
  title={{Normalization and Bias in Time Series Data}},
  booktitle={{Digital Interaction and Machine Intelligence}},
  year="2022",
  publisher="Springer International Publishing",
  address="Cham",
  pages="88--97",
  isbn="978-3-031-11432-8"
}

@inproceedings{wu2018group,
  title={{Group normalization}},
  author={Wu, Yuxin and He, Kaiming},
  booktitle={{Proceedings of the European conference on computer vision (ECCV)}},
  pages={3--19},
  year={2018}
}

@inproceedings{Wen_2021, 
  series={IJCAI-2021},
  title={{Time Series Data Augmentation for Deep Learning: A Survey}},
  doi={10.24963/ijcai.2021/631},
  booktitle={{Proceedings of the Thirtieth International Joint Conference on Artificial Intelligence}},
  publisher={{International Joint Conferences on Artificial Intelligence Organization}},
  author={Wen, Qingsong and Sun, Liang and Yang, Fan and Song, Xiaomin and Gao, Jingkun and Wang, Xue and Xu, Huan},
  year={2021},
  month=aug, pages={4653–4660},
  collection={IJCAI-2021} 
}

@misc{numpy2dhistogram,
  author = {{numpy.org}},
  title = {{numpy.histogram2d}},
  url = {https://numpy.org/doc/stable/reference/generated/numpy.histogram2d.html},
  year = {},
}

@inproceedings{inproceedingsEncodingTimeSeries,
  author = {Wang, Zhiguang and Oates, Tim},
  year = {2015},
  month = {01},
  pages = {},
  title = {{Encoding Time Series as Images for Visual Inspection and Classification Using Tiled Convolutional Neural Networks}}
}

@misc{pytorchSTFT,
  author = {{pytorch.org}},
  title = {{torch.stft}},
  url = {https://pytorch.org/docs/stable/generated/torch.stft.html},
  year = {},
}

@phdthesis{BerglerDissertation,
  title        = {{Deep Learning Applied to Animal Linguistics}},
  author       = {Christian Bergler},
  year         = 2023,
  month        = {9},
  note         = {Available at \url{https://nbn-resolving.org/urn:nbn:de:bvb:29-opus4-239268}},
  school       = {Friedrich-Alexander-Universität Erlangen-Nürnberg (FAU)},
  type         = {PhD thesis}
}

@misc{oord2018neuraldiscreterepresentationlearning,
  title={{Neural Discrete Representation Learning}}, 
  author={Aaron van den Oord and Oriol Vinyals and Koray Kavukcuoglu},
  year={2018},
  eprint={1711.00937},
  archivePrefix={arXiv},
  primaryClass={cs.LG},
  url={https://arxiv.org/abs/1711.00937}, 
}

@article{ILSVRC15,
  Author = {Olga Russakovsky and Jia Deng and Hao Su and Jonathan Krause and Sanjeev Satheesh and Sean Ma and Zhiheng Huang and Andrej Karpathy and Aditya Khosla and Michael Bernstein and Alexander C. Berg and Li Fei-Fei},
  Title = {{ImageNet Large Scale Visual Recognition Challenge}},
  Year = {2015},
  journal   = {International Journal of Computer Vision (IJCV)},
  doi = {10.1007/s11263-015-0816-y},
  volume={115},
  number={3},
  pages={211-252}
}

@article{van2008visualizing,
  title={{Visualizing data using t-SNE.}},
  author={Van der Maaten, Laurens and Hinton, Geoffrey},
  journal={Journal of machine learning research},
  volume={9},
  number={11},
  year={2008}
}

@inproceedings{Ekambaram_2023, 
  series={KDD ’23},
  title={{TSMixer: Lightweight MLP-Mixer Model for Multivariate Time Series Forecasting}},
  DOI={10.1145/3580305.3599533},
  booktitle={{Proceedings of the 29th ACM SIGKDD Conference on Knowledge Discovery and Data Mining}},
  publisher={ACM},
  author={Ekambaram, Vijay and Jati, Arindam and Nguyen, Nam and Sinthong, Phanwadee and Kalagnanam, Jayant},
  year={2023},
  month=aug, pages={459–469},
  collection={KDD ’23} 
}

@misc{wen2023transformerstimeseriessurvey,
  title={{Transformers in Time Series: A Survey}}, 
  author={Qingsong Wen and Tian Zhou and Chaoli Zhang and Weiqi Chen and Ziqing Ma and Junchi Yan and Liang Sun},
  year={2023},
  eprint={2202.07125},
  archivePrefix={arXiv},
  primaryClass={cs.LG},
  url={https://arxiv.org/abs/2202.07125}, 
}

@misc{cheng2024convtimenetdeephierarchicalfully,
  title={{ConvTimeNet: A Deep Hierarchical Fully Convolutional Model for Multivariate Time Series Analysis}}, 
  author={Mingyue Cheng and Jiqian Yang and Tingyue Pan and Qi Liu and Zhi Li},
  year={2024},
  eprint={2403.01493},
  archivePrefix={arXiv},
  primaryClass={cs.LG},
  url={https://arxiv.org/abs/2403.01493}, 
}

@misc{islam2023comprehensivesurveyapplicationstransformers,
  title={{A Comprehensive Survey on Applications of Transformers for Deep Learning Tasks}}, 
  author={Saidul Islam and Hanae Elmekki and Ahmed Elsebai and Jamal Bentahar and Najat Drawel and Gaith Rjoub and Witold Pedrycz},
  year={2023},
  eprint={2306.07303},
  archivePrefix={arXiv},
  primaryClass={cs.LG},
  url={https://arxiv.org/abs/2306.07303}, 
}

@misc{dosovitskiy2021imageworth16x16words,
  title={{An Image is Worth 16x16 Words: Transformers for Image Recognition at Scale}}, 
  author={Alexey Dosovitskiy and Lucas Beyer and Alexander Kolesnikov and Dirk Weissenborn and Xiaohua Zhai and Thomas Unterthiner and Mostafa Dehghani and Matthias Minderer and Georg Heigold and Sylvain Gelly and Jakob Uszkoreit and Neil Houlsby},
  year={2021},
  eprint={2010.11929},
  archivePrefix={arXiv},
  primaryClass={cs.CV},
  url={https://arxiv.org/abs/2010.11929}, 
}

@article{talukder2024totem,
  title={{Totem: Tokenized time series embeddings for general time series analysis}},
  author={Talukder, Sabera and Yue, Yisong and Gkioxari, Georgia},
  journal={arXiv preprint arXiv:2402.16412},
  year={2024}
}

@ARTICLE{10061349,
  author={Sun, Le and Liang, Jiancong and Zhang, Chunjiong and Wu, Di and Zhang, Yanchun},
  journal={IEEE Transactions on Intelligent Transportation Systems}, 
  title={{Meta-Transfer Metric Learning for Time Series Classification in 6G-Supported Intelligent Transportation Systems}}, 
  year={2024},
  volume={25},
  number={3},
  pages={2757-2767},
  doi={10.1109/TITS.2023.3250962}
}

@inproceedings{schuster25_interspeech,
  title     = {{PredTrAD – Prediction-based Transformer for Anomaly Detection in Multivariate Time Series Data}},
  author    = {Jan Schuster and Alexander Wölfel and Fabian Brunner and Christian Bergler},
  year      = {2025},
  booktitle = {{Interspeech 2025}},
  pages     = {3873--3877},
  doi       = {10.21437/Interspeech.2025-501},
  issn      = {2958-1796},
}

@misc{west2025pyscrewcomprehensivedatasetcollection,
  title={{PyScrew: A Comprehensive Dataset Collection from Industrial Screw Driving Experiments}}, 
  author={Nikolai West and Jochen Deuse},
  year={2025},
  eprint={2505.11925},
  archivePrefix={arXiv},
  primaryClass={cs.LG},
  url={https://arxiv.org/abs/2505.11925}, 
}

@online{WeberSchraubtechnik2025,
  author       = {{WEBER Schraubautomaten GmbH}},
  title        = {{Schraubtechnik Ratgeber}},
  year         = {2025},
  url          = {https://www.weber-online.com/schraubtechnik-ratgeber/}
}

@inproceedings{NIPS2017_3f5ee243,
  author = {Vaswani, Ashish and Shazeer, Noam and Parmar, Niki and Uszkoreit, Jakob and Jones, Llion and Gomez, Aidan N and Kaiser, \L ukasz and Polosukhin, Illia},
  booktitle = {{Advances in Neural Information Processing Systems}},
  editor = {I. Guyon and U. Von Luxburg and S. Bengio and H. Wallach and R. Fergus and S. Vishwanathan and R. Garnett},
  pages = {},
  publisher = {Curran Associates, Inc.},
  title = {{Attention is All you Need}},
  url = {https://proceedings.neurips.cc/paper_files/paper/2017/file/3f5ee243547dee91fbd053c1c4a845aa-Paper.pdf},
  volume = {30},
  year = {2017}
}

\end{document}